\documentclass[twocolumn]{article}
\usepackage{graphicx}
\usepackage{enumitem}
\usepackage{hyperref}
\usepackage{amsmath} 
\usepackage{amsfonts}
\usepackage{biblatex}
\usepackage{multicol}
\usepackage{mathtools,amssymb,lipsum}
\usepackage{cuted}
\usepackage{subcaption}
\usepackage{authblk}  

\title{Shaping Explanations: Semantic Reward Modeling with Encoder-Only Transformers for GRPO}
\author[1]{Francesco Pappone}
\author[2]{Ruggero Marino Lazzaroni}
\author[1]{Federico Califano}
\author[3]{Niccolò Gentile}
\author[4]{Roberto Marras}
\affil[1]{AI Sparks}
\affil[2]{University of Graz}
\affil[3]{Foyer Group}
\affil[4]{Onepix Academy}
\date{May 2025}
\bibliography{bib.bib}
\begin{document}
\twocolumn[
  \maketitle
  \begin{abstract}
    While Large Language Models (LLMs) excel at generating human-like text, aligning their outputs with complex, qualitative goals like pedagogical soundness remains a significant challenge. Standard reinforcement learning techniques often rely on slow and expensive LLM-as-a-judge evaluations or on brittle, keyword-based metrics like ROUGE \cite{lin-2004-rouge}, which fail to capture the semantic essence of a high-quality explanation. In this work, we introduce a novel approach to reward shaping within the Group Relative Policy Optimisation (GRPO) framework. Our central contribution is the use of a small, efficient encoder-only transformer as a semantic reward model. This model provides a dense, semantically rich reward signal based on the cosine similarity between a generated explanation and a ground-truth reference, guiding the policy towards explanations that are not just factually correct but also structurally and conceptually aligned with expert reasoning. We apply this method to the task of training a model for the Italian medical-school entrance examinations, following standard domain-adaptive continued pre-training (CPT) and supervised fine-tuning (SFT). Our results demonstrate that GRPO with our proposed semantic reward significantly improves explanation faithfulness and clarity over a strong SFT baseline, showcasing the power of using lightweight encoder models for nuanced reward shaping in complex generation tasks.
  \end{abstract}
  \vspace{1em}
]


\section{Introduction}
Frontier Large-Language Models (LLMs) such as the GPT series from OpenAI \cite{openai2023gpt4}, have begun to rival human experts on pedagogical tasks \cite{macina2025mathtutorbenchbenchmarkmeasuringopenended,gan2023largelanguagemodelseducation,shi2025educationqevaluatingllmsteaching}. While this success has sparked enthusiasm for using LLMs as automated tutors, generic models still remain optimized for broad helpfulness rather than rigorous pedagogy. Instruction-tuning with RLHF prioritises user satisfaction over didactic fidelity \cite{ouyang2022training, wen2024languagemodelslearnmislead}, and analyses of chain-of-thought faithfulness reveal that models often produce plausible, but logically untethered, explanations \cite{paul2024faithful}. In high-stakes contexts, such as in university entrance exams, these shortcomings are amplified by the need for step-by-step guidance that is both accurate and pedagogically sound.
The primary challenge lies in defining and optimizing for the quality of a generated explanation. Current reinforcement learning (RL) approaches face a trade-off. Using a powerful LLM-as-a-judge to score responses is computationally expensive, slow, and susceptible to biases like verbosity, and may be fragile to complex queries, especially when in less represented languages \cite{son2024llmasajudgerewardmodel}. Conversely, simpler reward functions based on lexical overlap (e.g., ROUGE \cite{lin-2004-rouge}) are efficient but shallow; they encourage keyword matching rather than a deep understanding of the target reasoning structure.
We address this gap by proposing a new reward shaping methodology for Group Relative Policy Optimisation (GRPO) \cite{shao2024grpo}. Our core idea is to leverage a small, efficient, encoder-only transformer as a specialized reward model. Instead of judging with natural language or counting keywords, this model computes dense vector embeddings of generated and reference explanations. The cosine similarity between these vectors serves as a powerful, semantically-grounded reward signal. This approach is computationally efficient, scalable, and captures the latent conceptual similarity that defines a high-quality explanation, all in a single forward pass of a small Transformer model. We demonstrate our method on a model trained for the Italian medical-school entrance exam, a domain requiring both specialized knowledge and clear, didactic rationales.

This work puts forward the following contributions.
\begin{itemize}
    \item A novel reward-shaping framework for GRPO that uses an efficient encoder-only model to provide a dense, semantic similarity-based reward for explanation quality.
    \item A demonstration that this semantic-based reward, combined with auxiliary rewards for correctness and formatting, significantly improves explanation faithfulness and clarity over strong SFT baselines.
    \item An empirical analysis showing that the resulting model outperforms its SFT counterpart not only on the target domain but also on out-of-domain reasoning tasks, as evaluated by an external LLM judge.
    \item As an ablation study, motivated by \cite{pascanu2025optimizersqualitativelyaltersolutions},  we study the effect in the use of the Muon optimizer \cite{jordan2024muon} CPT. We show it does not provide significant gains for our case. 
\end{itemize}
 Section \ref{sec:background} surveys related efforts in LLM alignment and reward modeling. Section \ref{sec:method} describes our  pipeline, focusing on the GRPO stage and our encoder-based reward design. Section \ref{sec:experiment} details the Experiments, with Section \ref{sec:data_preparation} in particular focusing on corpus curation. Results and ablations appear in Section \ref{sec:results}, followed by discussion in Section \ref{sec:discussion}.

\section{Background and Related Work}
\label{sec:background}
\paragraph{Reward models and explanation faithfulness.} Preference‑based fine‑tuning such as RLHF can inadvertently incentivise \emph{plausible but unfaithful} \cite{Huang_2025} rationales. Ferreira \emph{et~al.} enrich reward‑model inputs with causal‑attribution signals and empirically curb reward hacking in explanations~\cite{ferreira2025automatic}. Paul \emph{et~al.} frame rationale faithfulness as a causal mediation problem and propose the \textsc{FRODO} framework to train models whose reasoning genuinely supports their answers~\cite{paul2024faithful}.

\paragraph{RL algorithms for reasoning‑centric objectives.} To reduce memory overhead while still benefiting from policy gradient updates, Shao \emph{et~al.} introduce \textsc{GRPO} as part of \textsc{DeepSeekMath}; the algorithm compares a batch of candidate responses against their group‑mean baseline and delivers sizeable gains on math benchmarks~\cite{shao2024deepseekmathpushinglimitsmathematical}. GRPO has become the de-facto standard in modern post-training of models, mainly in the form of Reinforcement Learning with Verifiable Rewards (RLVR), starting from DeepSeek R1's success in producing an open source LLM that is capable of reasoning \cite{deepseekai2025deepseekr1incentivizingreasoningcapability}.  

\paragraph{Encoder-only reward models.}
Light-weight encoder transformers have increasingly been adopted as reward models since they are cheaper to run than LLM judges yet capture rich semantics.  For example, \textsc{MisinfoCorrect} uses a BERT classifier to reward factual, polite counterspeech~\cite{he2023misinfocorrect}.  In abstractive summarisation, BERTScore embeddings supply a distributional-semantic reward that outperforms ROUGE in RL fine-tuning~\cite{li2019dsr}.  Most relevant to our work, Li~\textit{et al.} integrate a frozen RoBERTa-base multi-label regressor directly into the GRPO loop, demonstrating that encoder-only rewards scale cleanly with group-normalised policy optimisation~\cite{li2025multiobjectivegrpo}. Differing from our case, they use the RoBERTa model to output a series of scores, while we directly use cosine similarity with respect to the generated embedding.

\paragraph{Self‑reward and pedagogical alignment.} Chen \emph{et~al.} show that LLMs can bootstrap hidden reasoning skills via a self‑rewarding variational formulation (\textsc{LaTRO}) without external feedback~\cite{chen2024latro}. Complementarily, Sonkar \emph{et~al.} collect human preferences for \emph{pedagogical alignment}, tuning models that guide learners through sub‑questions rather than revealing answers directly~\cite{sonkar2024pedagogical}. Our work shares this tutoring objective while focusing on high‑stakes medical‑school admission content.

\vspace{0.5em}
Compared with the above literature, our primary innovation is the use of a lightweight encoder-only transformer as a semantic similarity scorer within the GRPO framework, providing a practical and effective alternative to both LLM-as-a-judge and more naive lexical-overlap rewards for aligning explanations.

\section{Methodology}
\label{sec:method}
Our training pipeline consists of three stages: domain-adaptive continued pre-training (CPT), supervised fine-tuning (SFT) to instill a task-specific format, and a final reinforcement learning stage using GRPO, where our novel reward shaping is applied. We train starting from an Italian fine-tune of DeepSeek-R1-Distill-Llama-8b \cite{deepseekai2025deepseekr1incentivizingreasoningcapability}.

\subsection{Stage~I: Continued Pre‑training (CPT)}
To equip the model with specialized knowledge, we perform Continued Pre-training over a 4M-token textbook corpus using a causal language modeling objective. This stage uses full-precision parameter updates, a sequence length of 4,096 tokens, and a standard linear warmup, cosine decay learning rate schedule. CPT is crucial for grounding the model in the target domain. We train the model for a total of 5 epochs on the dataset. 
\paragraph{Muon Optimizer Ablation.} For this stage, we conducted an ablation comparing the standard AdamW \cite{loshchilov2019decoupledweightdecayregularization} optimizer against the more novel Muon optimizer \cite{jordan2024muon}. Muon is designed to improve training stability and efficiency for matrix-valued parameters. It augments standard SGD-with-momentum updates with a post-processing step that performs a matrix semi-orthogonalization, which may also be seen as steepest descent under the spectral norm \cite{cesista2025sdnr}. Empirically, update matrices in transformers often exhibit a high condition number, meaning the updates are dominated by a few directions. Orthogonalization is hypothesized to amplify the signal from rarer but still important learning directions. Muon achieves this using the Newton-Schulz (NS) iteration,  effectively finding the nearest semi-orthogonal matrix to the original update. This technique is only used to orthogonalize updates to matrix parameters, while non-matrix ones (such as biases) remain optimized using regular AdamW. We use Moonlight's \cite{liu2025muonscalablellmtraining} implementation, designed to match AdamW's learning rate.
\subsection{Stage~II: Supervised Fine‑Tuning (SFT)}
Next, we fine-tune the previous continuously pre-trained model on our question-and-explanation dataset to teach it the desired output format. Each example follows a structured template, and the model is trained via token-level cross-entropy to predict both the correct answer and the detailed rationale. This section efficiently warms the model to our Q\&A (with explanation) data format, and significantly improves our benchmark results.

\subsection{Stage~III: GRPO with Semantic Reward Shaping}
\paragraph{Motivation.} Current methods for generating reward signals in unverifiable domains fall into two categories. The first, "LLM-as-a-judge", uses large, multi-forward pass and computationally expensive models to provide numerical or probabilistic scores, but these can often be biased, brittle, and even miscalibrated. The second, keyword-based rewards like ROUGE \cite{lin-2004-rouge}, is computationally cheaper but fails to reward semantic understanding, instead encouraging models to simply memorize key terms from the golden response. Our work introduces a third path that provides both a semantic understanding of the target, as well as relative computational efficiency. We use small, efficient encoder-only models to create a semantically rich reward signal. By computing the cosine similarity between the generated and target embeddings in a single forward pass, we directly incentivize the model to learn abstract skills, such as high-quality explanation, rather than just mimicking surface-level features.

\subsection{Reward Model Design}
To align our model towards generating high-quality, structured, and accurate explanations, we employ a multi-component reward function within a Group Relative Policy Optimization (GRPO)
framework~\cite{shao2024deepseekmathpushinglimitsmathematical,deepseekai2025deepseekr1incentivizingreasoningcapability}. Starting from a continued pretrained and Supervised Fine-Tuned (SFT) checkpoint, we perform 1,000 gradient steps of GRPO with a group size of $K=6$ generations per prompt. Each generation is scored by the sum of four distinct reward signals detailed below.

\paragraph{Reward Components.}
The total reward $R$ for a given generation is a sum of the following four components:
\begin{enumerate}[nosep, leftmargin=*]
    \item  Semantic Similarity, the core of our reward framework. This component measures the conceptual and structural alignment between the generated explanation and the ground-truth reference. It is calculated as the cosine similarity of their respective dense vector embeddings, which are produced by a pre-trained 600M-parameter encoder-only transformer (qwen3-0.6B \cite{qwen3}). A detailed explanation is provided in the following section. This is swapped for a ROUGE-L reward or an LLM-as-a-judge reward for later experiments. (\emph{Semantic+ROUGE GRPO} adds the ROUGE-L F1 component to the same unweighted sum.)

    \item  A binary reward of 1.0 if the model's final answer, extracted from the \texttt{<risposta>} tag, exactly matches the ground-truth answer, and 0.0 otherwise. This ensures factual accuracy.
    
    \item  A rule-based reward that grants a score of 1.0 only if the output correctly uses the required XML tags (\texttt{<spiegazione>} and \texttt{<risposta>}) to structure the output. This promotes structural predictability.
    
    \item  A reward for including a non-empty "chain-of-thought" block within the designated \texttt{<think>} tags, encouraging the model to externalize its reasoning process.
\end{enumerate}

\paragraph{Semantic Similarity Reward Calculation.}
The primary reward signal is designed to move beyond simple lexical overlap (e.g., ROUGE \cite{lin-2004-rouge}) and capture deeper semantic meaning. Let $\mathbf{v}_{\text{gen}} \in \mathbb{R}^d$ be the dense embedding of the model-generated explanation and $\mathbf{v}_{\text{gt}} \in \mathbb{R}^d$ be the embedding of the ground-truth reference, where $d$ is the dimensionality of the encoder's embedding space. The reward, $R_{\text{sim}}$, is their adjusted cosine similarity, defined as:
\[
R_{\text{sim}} = \cos(\theta) - \cos(\theta_{ref}) = \frac{\mathbf{v}_{\text{gen}} \cdot \mathbf{v}_{\text{gt}}}{\|\mathbf{v}_{\text{gen}}\| \|\mathbf{v}_{\text{gt}}\|} - \frac{\mathbf{v}_{\text{gen}} \cdot \mathbf{v}_{\text{ref}}}{\|\mathbf{v}_{\text{gen}}\| \|\mathbf{v}_{\text{ref}}\|}
\]
where $\|\cdot\|$ denotes the Euclidean (L2) norm. Since the reward is a difference of cosine similarities, we rescale the result by a coefficient of $c=4$, so that the reward tends to sit in the 0 to 1 range. We find that the reward is never negative in practice and spans well the 0-1 range.
The $\cos(\theta_{ref})$ term is obtained by embedding a set of random vectors in the dataset and taking the average embedding $\mathbf{v}_{ref}$. This is necessary due to cosine similarity being generally high among explanation vectors. 
In our implementation, we obtain these vectors by passing each text through the encoder model. This process yields a single fixed-size vector for each explanation.  

\paragraph{GRPO settings.} 

We train a LoRA \cite{hu2021loralowrankadaptationlarge} adapter with rank $r = 32$ and $\alpha = 64$, using a clipping parameter $\varepsilon=0.2$ and KL coefficient $\beta=0.05$. The KL divergence term $D_{\text{KL}}$ penalizes large deviations from the original policy to stabilize training. We use a generation temperature of 0.7. We found these coefficients to work well across a series of different tested hyperparametric configurations. 

\section{Experiment}
\label{sec:experiment}
\subsection{Italian admission test primer}
The medicine entrance exam comprises multiple-choice questions in biology, chemistry, physics, mathematics, and logic.
Items test conceptual understanding and applied reasoning under strict time limits, making concise, didactic explanations
particularly valuable.

\subsection{Data Preparation}
\label{sec:data_preparation}
We assemble and preprocess two complementary corpora: a curated textbook collection for continued pre‐training, and a detailed question–answer–rationale set for supervised fine‐tuning and reward‐shaping.

\subsubsection{Textbook Corpus Construction}
Our textbook corpus comprises twenty‐eight Italian volumes covering biology, chemistry, physics, mathematics, and logic, provided by OnePix Academy S.R.L., supplemented by open‐access lecture notes. In total, the raw PDFs yield approximately 4 million tokens after processing. We first apply OCR correction to repair common character‐recognition errors, then remove headers, footers, page numbers, and other formatting artifacts, all while preserving the original chapter and section structure. To ensure diversity and avoid redundancy, we deduplicate at the paragraph level using exact‐match and near‐duplicate detection. Finally, each document is tokenized with a LLaMA 3 tokenizer \cite{grattafiori2024llama3herdmodels} and split into overlapping windows of up to 4096 tokens, with metadata recording the source title and chunk position. This results in two published variants of the textbook corpus: the full‐text version and a chunked version optimized for long‐context training.

\subsubsection{Question–Answer–Rationale Sets}
The question–answer dataset contains 19014 multiple‐choice items from Italian medical‐school entrance exams (2011–2024), each paired with a tutor‐crafted, stepwise explanation. We clean the source files by stripping HTML tags and filtering out any items that reference external images. Metadata, such as subject, main topic, and subtopic, are automatically extracted from filenames and headers.
The resulting collection is stratified by subject and difficulty (levels 1–3) and split into 80 \% train, 10 \% development, and 10 \% blind‐test subsets. Building on this foundation, we derive two instruction-tuning variants by converting each question into a concise instruction prompt with its corresponding answer and explanation, filtering out entries with missing or overly brief rationales.

\subsection{Evaluation Framework}
\subsubsection{LLM-Judged Cross-Benchmark Comparison}
\label{sec:evaluation}
To probe whether the explanation gains observed on admission-exam items
translate to \emph{out-of-domain reasoning}, we run a head-to-head test on the
\texttt{minibench-reasoning} benchmark, a set of reasoning-intensive questions from MedBench-ita \cite{lazzaroni2025medbenchitcomprehensivebenchmarkevaluating}, which is a benchmark built to evaluate LLMs on Italian language multiple-choice questions tailored for the Italian Medicine University entrance exam. The subset we used features questions on Logic, Mathematics, and Physics. For each question we pair the explanations produced by the checkpoints and present them, anonymised, to an external judge model.  The judge selects the superior answer based on logical soundness, clarity, completeness, and focus, or records a tie when neither is clearly better.


\section{Results \& Analysis}
\label{sec:results}

\begin{figure}[t]
  \centering
  \includegraphics[width=\linewidth]{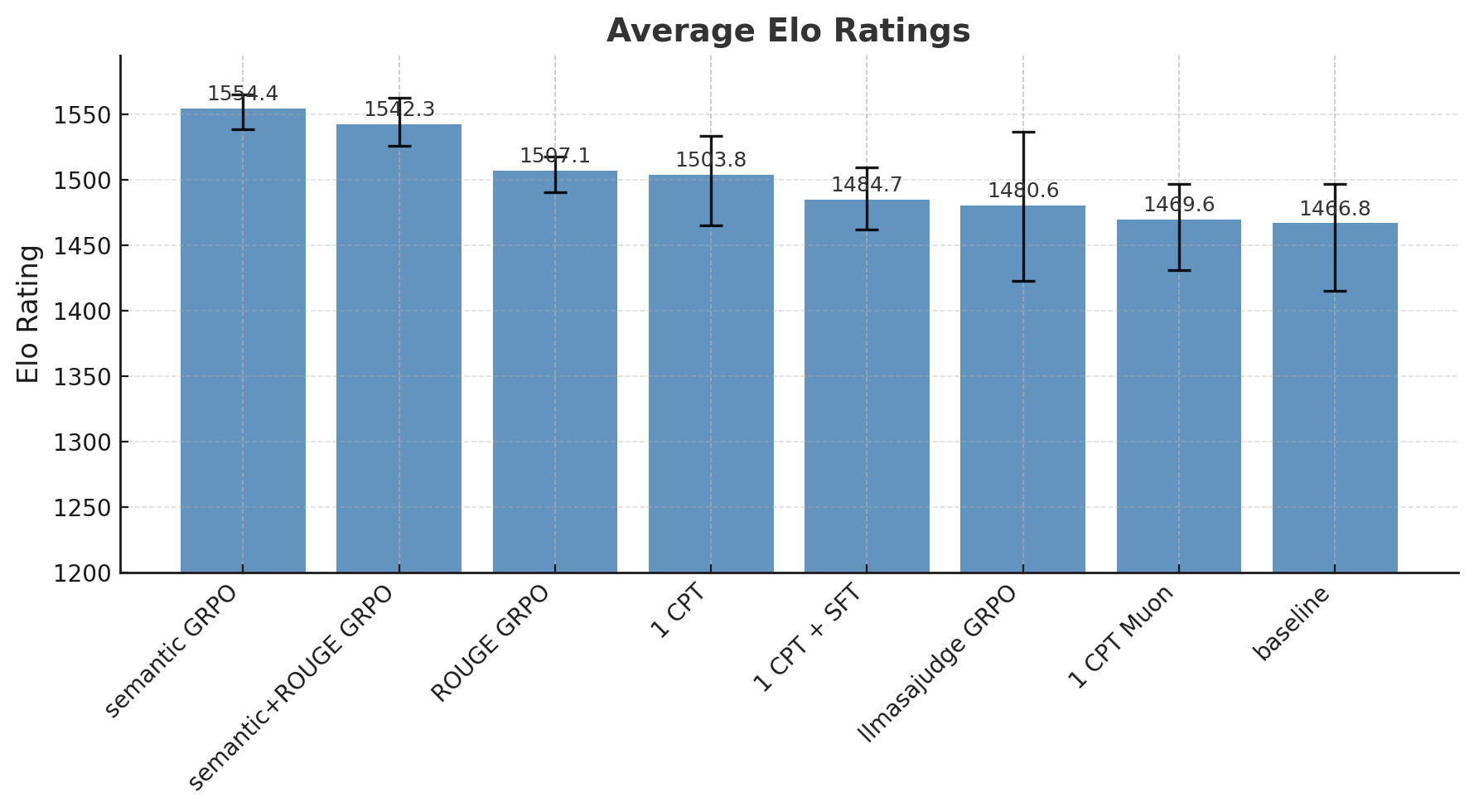} 
  \caption{\textbf{Average Elo ratings across model variants.} Error bars show the range across three different judge models.}
  \label{fig:avg_elo}
\end{figure}

\paragraph{Setup.} We report Elo ratings computed from pairwise, anonymised comparisons of model outputs on held-out items, judged by three external LLM as in \S\ref{sec:evaluation}, namely gpt-oss 20b \cite{openai2025gptoss120bgptoss20bmodel},  GPT-5-nano and DeepSeek R1 Qwen3 8B \cite{deepseekai2025deepseekr1incentivizingreasoningcapability}. Higher is better. Each variant is evaluated across three independent runs for sake of self-consistency \cite{wang2023selfconsistencyimproveschainthought}; error bars in Fig.~\ref{fig:avg_elo} denote the min–max range across runs.

\paragraph{Main finding.} \textbf{GRPO with the semantic similarity reward (Semantic GRPO) is the top performer.} The \emph{semantic GRPO} variant attains the highest Elo (1554.4), outperforming all alternatives by a clear margin: +12.1 over \emph{semantic+ROUGE GRPO} (1542.3), +47.3 over \emph{ROUGE GRPO} (1507.1), and +87.6 over the \emph{baseline} (1466.8). Furthermore, in a specialized comparison between RL and classical methods, we see that 
\textbf{Semantic GRPO is the top performer in both accuracy and preference in explanation.}
The \emph{semantic} variant attains the highest Elo (1574.4) and the best reasoning accuracy (0.581), narrowly ahead of ROUGE (0.580) and LLM-as-judge (0.580), while substantially outperforming all non-RL posttraining baselines in preference.

\paragraph{Reward-design ablations.}
\begin{itemize}[leftmargin=*,nosep]
  \item \textbf{Adding ROUGE slightly hurts the semantic signal.} Combining the semantic reward with ROUGE yields 1542.3 Elo, trailing the pure semantic variant by 12.1 points. This suggests lexical-overlap pressure can dilute the intended semantic alignment.
  \item \textbf{ROUGE-only improves over baselines but lags semantic rewards.} \emph{ROUGE GRPO} reaches 1507.1, comfortably above \emph{baseline} (+40.3) yet far from \emph{semantic GRPO} (–47.3).
  \item \textbf{LLM-as-judge GRPO is less competitive and more variable.} The \emph{LLM-judge GRPO} variant scores 1480.6 with visibly larger run-to-run spread, indicating instability and lower average quality relative to embedding-based rewards.
\end{itemize}

\paragraph{Pre-training and SFT baselines.}
\begin{itemize}[leftmargin=*,nosep]
  \item \textbf{CPT helps; SFT alone does not close the gap.} A single epoch of CPT (\emph{1 CPT}) yields 1503.8 Elo. Adding SFT without RL (\emph{1 CPT + SFT}) drops to 1484.7 (–19.1 vs.\ CPT), indicating that format-conditioning alone does not guarantee stronger judged preferences.
  \item \textbf{Optimizer ablation (CPT).} \emph{1 CPT Muon} reaches 1469.6, roughly on par with \emph{baseline} (+2.8) and substantially below \emph{1 CPT} with AdamW (–34.2). In our configuration, Muon did not translate into higher judged quality at this stage. This is in line with recent literature showing how matrix-based optimizers may provide decreasing benefits to increasingly large language models \cite{wen2025fantasticpretrainingoptimizers}.
\end{itemize}

\paragraph{Takeaways.} (i) The semantic encoder reward is the key driver of judged quality; (ii) mixing lexical rewards with semantics is counter-productive for our task; (iii) CPT provides a useful foundation, and SFT alone does not match the gains from RL with a semantic reward; and (iv) LLM-as-a-judge rewards underperform and are less stable than lightweight encoder rewards in this setup.

\begin{figure}[t]
  \centering
  \includegraphics[width=\linewidth]{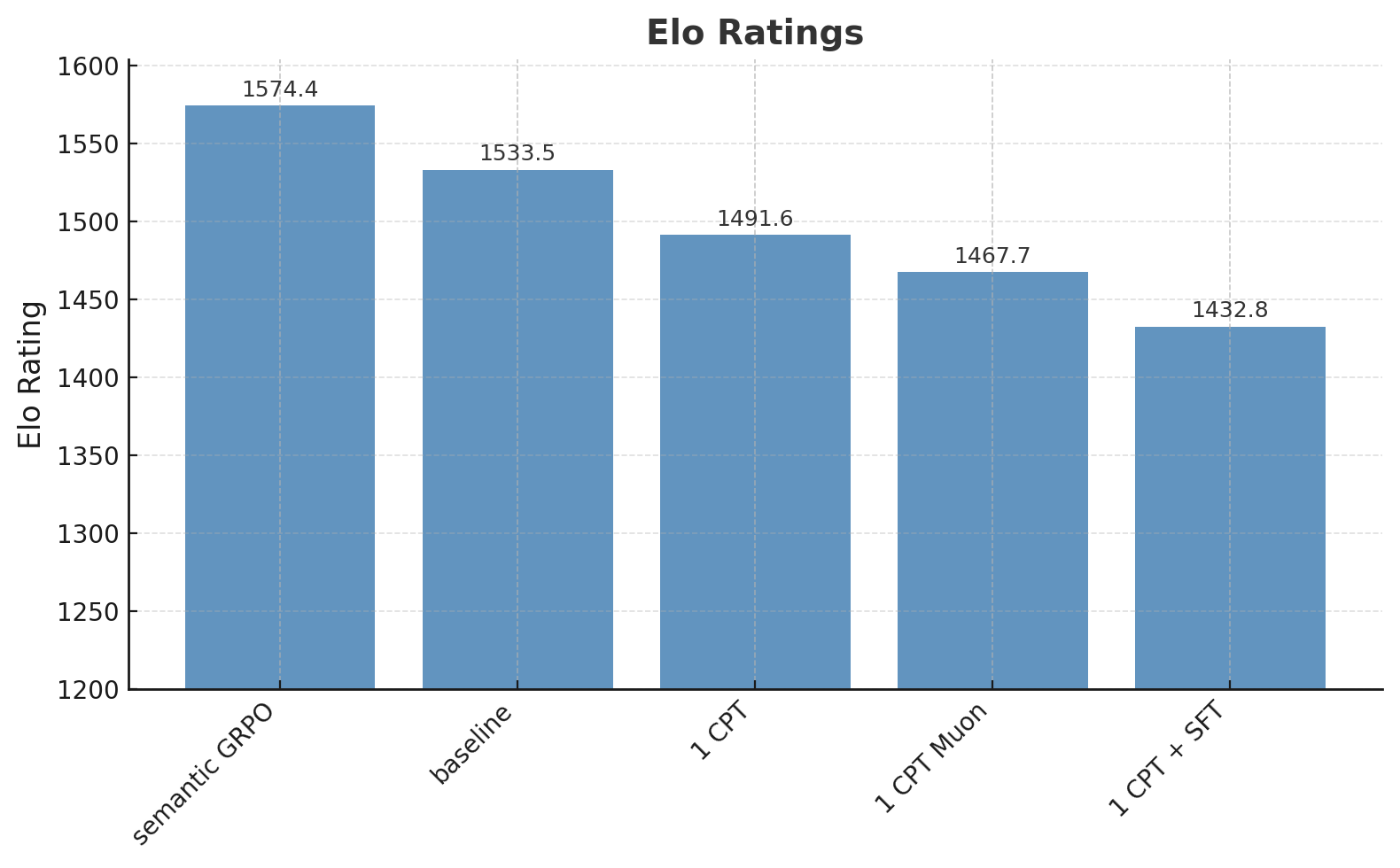} 
  \caption{Elo ratings of models trained with different combinations of continued pretraining (CPT) and supervised fine-tuning (SFT), evaluated by GPT-5-nano.}
  \label{fig:elo_final}
\end{figure}

\begin{figure}[t]
  \centering
  \includegraphics[width=\linewidth]{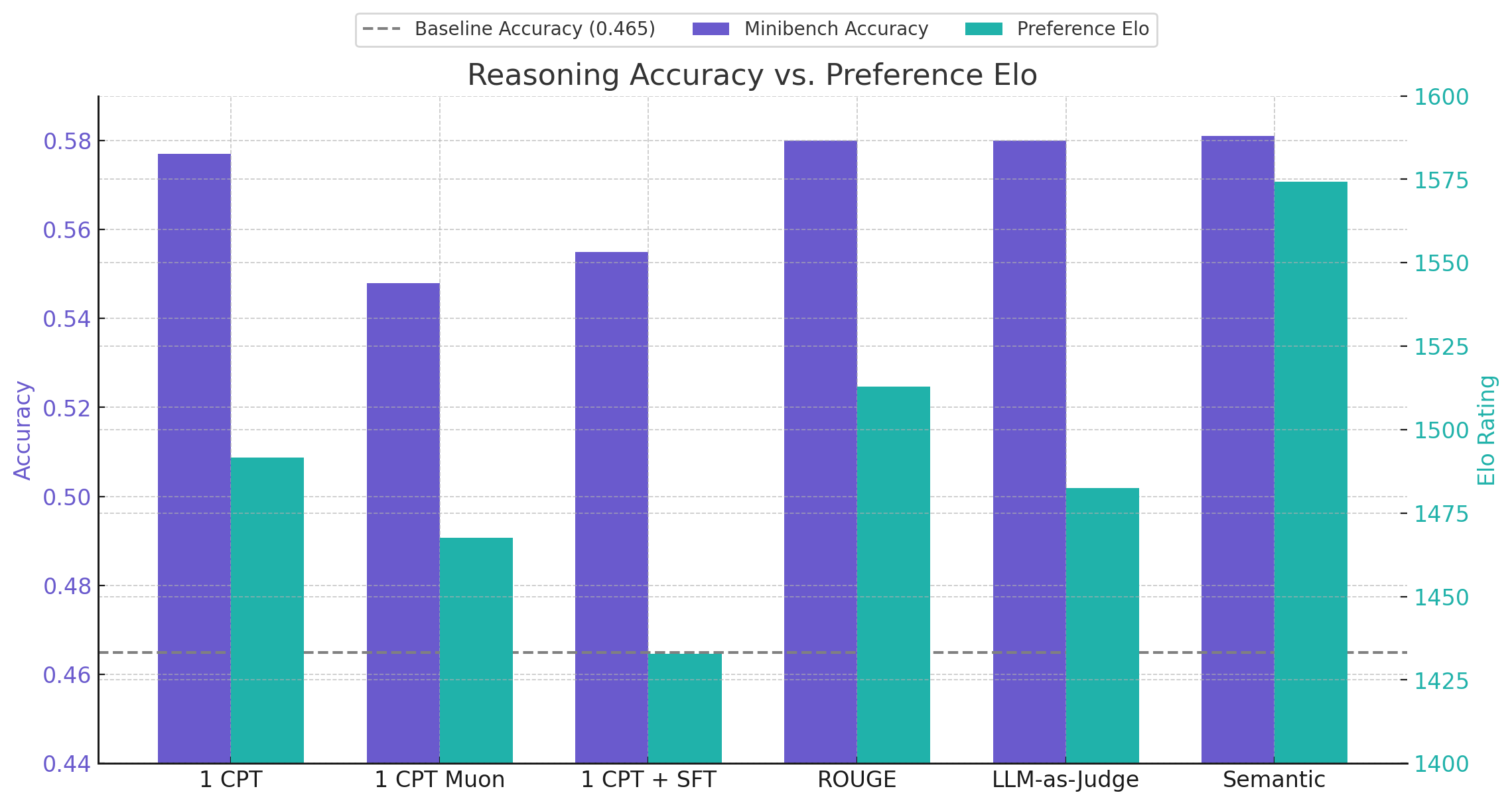}
  \caption{\textbf{Reasoning accuracy vs.\ preference Elo.}
  Bars (left axis) report \texttt{minibench-reasoning} accuracy; bars (right axis) report Elo from LLM-judge preferences. The dashed line marks the baseline accuracy (0.465). Semantic reward alignment is the only setting that improves both accuracy and preference simultaneously.}
  \label{fig:elo-vs-acc}
\end{figure}

\paragraph{Posttraining and memorization.} We conducted a focused comparison using GPT-5-nano as a judge to probe the effects of continued pretraining and fine-tuning on explanation quality. The results (Fig.~\ref{fig:elo_final}) suggest that while regular posttraining strategies like CPT and SFT can increase memorization and task format adherence, they do not necessarily improve—and can even degrade—the conceptual quality of explanations. In contrast, the \emph{semantic GRPO} model significantly outperforms all variants lacking reinforcement alignment, indicating that reward-guided fine-tuning plays a crucial role in achieving high-quality pedagogical outputs.
\paragraph{Accuracy vs.\ preference (complementary view).}
Fig.~\ref{fig:elo-vs-acc} complements the Elo-only analyses by jointly plotting reasoning accuracy and preference.
While several posttraining routes reach similar accuracy (\(\approx 0.58\)), only the \emph{Semantic} variant
also yields a clearly higher Elo (1574.4). In contrast, \emph{1 CPT} and \emph{1 CPT + SFT} raise accuracy relative to the
baseline (0.465) but do not translate these gains into preferred explanations, aligning with the trends in
Figs.~\ref{fig:avg_elo} and \ref{fig:elo_final}.
\section{Discussion}
\label{sec:discussion}
\begin{itemize}[nosep]
    \item \textbf{The power of semantic reward.} Our results clearly show that while CPT contributes domain knowledge and SFT imparts task structure, the crucial step in achieving high-quality explanations is the GRPO phase. The encoder-based semantic reward successfully guides the model to produce rationales that are conceptually aligned with expert solutions, moving beyond surface-level correctness.
    \item \textbf{Efficiency and Scalability.} Using a lightweight encoder model for reward computation is orders of magnitude faster and cheaper than using an LLM-as-a-judge, making this approach highly practical for iterative development and larger-scale training runs.
    \item \textbf{Limitations.} The quality of the semantic reward is bounded by the capabilities of the encoder model and the quality of the reference explanations. While effective, the reward model could still be gamed in unforeseen ways.
    We intentionally used an unweighted sum of components. While this keeps the setup simple and reproducible,
it also means components with different numerical ranges can influence the total reward unevenly.
A principled normalisation/weighting scheme is promising future work.
    
\end{itemize}

\section{Conclusion \& Future Work}
We have presented a novel and effective method for improving LLM explanation quality using Group Relative Policy Optimisation. By replacing expensive or brittle reward functions with a lightweight, encoder-only transformer that provides a dense semantic similarity score, we successfully aligned a model with the complex goal of generating pedagogically sound explanations for the Italian medical-school entrance exam. This work demonstrates that small, specialized models can play a crucial role in shaping the behavior of their larger counterparts in a targeted and efficient manner.
Future work will explore applying this technique to multi-turn tutoring dialogues, and investigating its transferability to other languages and domains.

\paragraph{Acknowledgements}
We acknowledge this research was commissioned  by OnePix Academy SRL, with the objective of improving the quality of the explanation of their production models. 
\printbibliography
\end{document}